\documentclass[runningheads]{llncs}
\usepackage{graphicx}
\usepackage{amsmath,amssymb} 
\usepackage{color}
\usepackage[width=122mm,left=12mm,paperwidth=146mm,height=193mm,top=12mm,paperheight=217mm]{geometry}
\begin{document}
\pagestyle{headings}
\mainmatter

\title{Generating captions without looking\\beyond objects}

\titlerunning{Generating captions without looking beyond objects}

\authorrunning{H. Heuer, C. Monz, A.W.M. Smeulders}

\author{Hendrik Heuer\textsuperscript{1}, Christof Monz\textsuperscript{2}, Arnold W.M. Smeulders\textsuperscript{1}}
\institute{\textsuperscript{1}QUVA Lab, \textsuperscript{2}Informatics Institute,\\
	University of Amsterdam, The Netherlands\\
	\email{ \{h.heuer,c.monz,a.w.m.smeulders\}@uva.nl}
}

\maketitle

\begin{abstract}
This paper explores new evaluation perspectives for image captioning and introduces a noun translation task that achieves comparative image caption generation performance by translating from a set of nouns to captions. This implies that in image captioning, all word categories other than nouns can be evoked by a powerful language model without sacrificing performance on n-gram precision. The paper also investigates lower and upper bounds of how much individual word categories in the captions contribute to the final BLEU score. A large possible improvement exists for nouns, verbs, and prepositions. 
\end{abstract}

\section{Introduction}

\begin{figure}[t]
\begin{center}
\includegraphics[height=3cm]{Figures/dog_tug_of_war_better.png}
\end{center}
   \caption{``Woman and dog with frisbee on grass near fence.'' and ``a woman playing tug of war with a dog over a white frisbee.'' are both valid description of this image.}
\end{figure}

The objective of image captioning is to automatically generate grammatical descriptions of images that represent the meaning of a single image. Figure 1 compares different captions for an image from the MSCOCO \cite{DBLP:journals/corr/ChenFLVGDZ15} dataset. Describing Figure 1 as ``woman and dog with frisbee on grass near fence.'' is merely turning the results of an object detection into a fluent English sentences. Generating a caption such as ``a woman playing tug of war with a dog over a white frisbee.'' exhibits an understanding that the dog is biting a frisbee, that the woman is playing with the dog and, more specifically, that the woman and the dog are playing a game called tug of war. Captions like this are more informative and relevant. They could also be beneficial in an assistive technology context where captions help visually impaired people, e.g. on Facebook. 

This paper motivates why it is important to dissociate the captioning task from the object detection task \`{a} la ImageNet and why new evaluation perspectives are needed. For this, it introduces a blind noun translation task that compares an image captioning system to the intrinsic language modeling capabilities of state-of-the-art recurrent neural networks. The paper also analyzes the image captions generated by three state-of-the-art systems: the Karpathy et al. \cite{karpathy2015deep} system, the Vinyals et al. \cite{DBLP:journals/corr/VinyalsTBE14} system, and the Xu et al. \cite{DBLP:journals/corr/XuBKCCSZB15} system. The systems are analyzed in regards to the contribution of different word categories, i.e., part of speech tags, to the BLEU score.

\section{Background}

Recent approaches in automated caption creation have addressed the problem of generating grammatical descriptions of images as a sequence modeling problem using recurrent neural network (RNN) language models \cite{karpathy2015deep,DBLP:journals/corr/VinyalsTBE14,DBLP:journals/corr/XuBKCCSZB15,DBLP:journals/corr/WuSHLD15}.

Both Karpathy et al. \cite{karpathy2015deep} and Vinyals et al. \cite{DBLP:journals/corr/VinyalsTBE14} combine a vision CNN with a language generating RNN. For each image, a feature vector of the last fully-connected layer of a CNN pretrained on ImageNet is fed into an RNN. Xu et al. \cite{DBLP:journals/corr/XuBKCCSZB15} extended this architecture by adding an attention mechanism, which learns not only a distribution over the words in the vocabulary but also a distribution over the locations in the image based on the last convolutional layer of a CNN pretrained on ImageNet. Wu et al. \cite{DBLP:journals/corr/WuSHLD15} introduced a method of incorporating explicit high-level concepts such as bag, eating, and red, which is remarkable as it covers a noun, a verb, and an adjective. Their attribute-based V2L framework consists of an image analysis module that learns a mapping between an image and the semantic attributes through a CNN, as well as a language module, that learns a mapping from the attributes vector to a sequence of words using an LSTM. The suggested broad coverage over language is misleading, however. Nouns are well covered these days by concept detectors. ``Eating'' is a visually well-defined state descriptor which can generally be captured in an attribute. This does not hold for many other actions. ``Red'' is likewise well described. It is one of the rare adjectives to have that property. In the sequel, we will argue that current error measures like BLEU are too insensitive to capture this. This work is complementary to previous work that disentangled the contribution of visual model and language model \cite{DBLP_journals_corr_YaoBCSB15}.

\begin{figure}[t]
\begin{center}
\includegraphics[height=4cm]{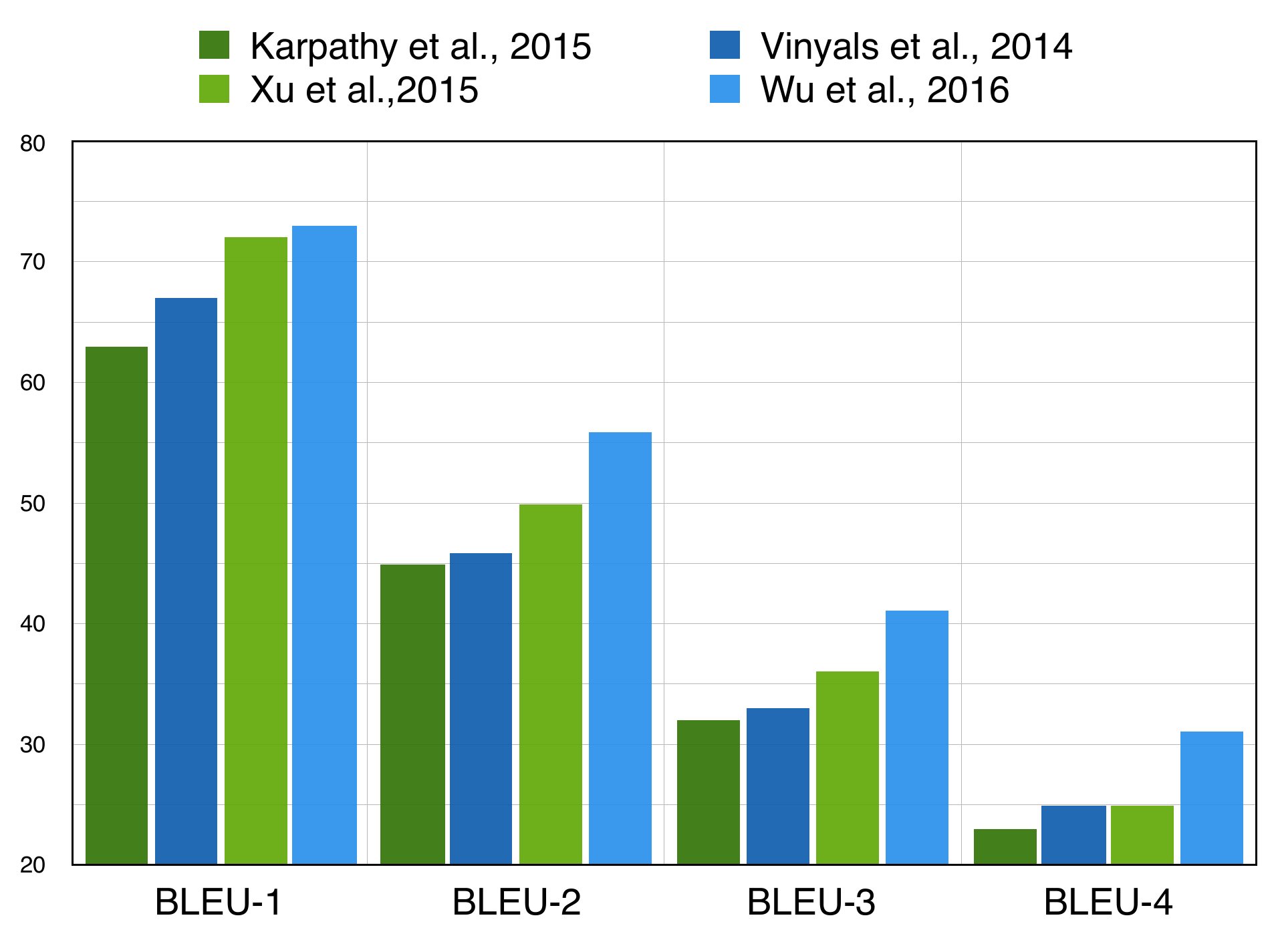}
\end{center}
   \caption{Comparison of state-of-the-art image captioning systems on the MSCOCO 2014 Captioning Challenge Testing data regarding the BLEU metric (higher scores indicate better precision performance).}
\end{figure}

BLEU is a precision-oriented machine translation evaluation metric that measures the N-gram overlap between sentences in the output of a machine translation system and one or more reference translations \cite{Papineni:2002:BMA:1073083.1073135}. The clipped n-gram counts for all the candidate sentences are divided by the number of candidate n-grams in the test corpus to compute a precision score for the entire test corpus. BLEU ignores the relevance of words and only operates on a local level without taking the overall grammaticality of the sentence or the sentence meaning into account. 

Figure 2 compares four state-of-the-art models regarding their BLEU-1-4 performances. While a significant improvement over the last years can be observed, it is not clear how much individual word categories, i.e., part-of-speech tags in the captions, contribute to the final BLEU score.

\section{Methodology}

We introduce a blind noun translation task that demonstrates the language modeling capabilities of state-of-the-art language models like LSTMs and practically shows how much a language model can infer from a set of nouns. For the blind noun translation task, a state-of-the-art neural machine translation model \cite{DBLP:journals/corr/BahdanauCB14} is trained to translate nouns into full captions. These captions are compared to the Karpathy et al. \cite{karpathy2015deep} image captioning system. The nouns fed to the translation model are not taken from the ground truth of the dataset but are automatically extracted from the captions generated by the Karpathy et al. system to increase comparability of the results. For this, the Stanford log-linear part-of-speech tagger \cite{Toutanova:2000:EKS:1117794.1117802,Toutanova03feature-richpart-of-speech} is used to extract all noun tags. 

The model for the blind noun translation task is trained on the MSCOCO 2014 \cite{DBLP:journals/corr/ChenFLVGDZ15} training set, which includes over 80,000 images with 5 captions per image. The neural machine translation model was trained on 500,000 training pairs, which each consists of a caption on the target side and a random permutation of the nouns in the caption on the source side. Pairs of nouns such as ``plate food table woman'' are translated into target sentences such as ``A woman sitting at a table eating a plate of food''. For the evaluation of the blind noun translation experiment, the MSCOCO 2014 \cite{DBLP:journals/corr/ChenFLVGDZ15} validation set is used, which consists of more than 40,000 captions. 

To compare the contribution of individual word categories, the generated output of three state-of-the-art systems from Karpathy et al. \cite{karpathy2015deep}, Vinyals et al. \cite{DBLP:journals/corr/VinyalsTBE14}, and Xu et al. system \cite{DBLP:journals/corr/XuBKCCSZB15} was obtained and analyzed. Since different subsets of the datasets were provided, the results are not directly comparable with one another. That said, they give an indication of the contribution of specific word categories. 

For each system, a set of generated captions was selected. Using the Stanford log-linear part-of-speech tagger \cite{Toutanova:2000:EKS:1117794.1117802,Toutanova03feature-richpart-of-speech}, for each word in the caption, the word category based on its syntactic function was assigned (noun, verb, adjective). Using this, the lower and the upper bound were determined. For both approaches, a certain word category was replaced by a special token. For the upper bound, only the verbs in the generated captions are replaced by this token, meaning that, effectively, no verb is correct, since they will always be different from the references. For the lower bound, all verbs in both the references and the generated captions are replaced by this token, which means that every verb is detected correctly.

\section{Results of the blind noun (theoretical) experiment}

Table 1 shows the performance on the blind noun translation task in comparison to the n-gram precision scores of the Karpathy et al. image captioning system. The difference in performance according to the n-gram precision is relatively small, ranging from only 1.1\% for 4-grams up to 3.2\% for bigrams. This indicates that with respect to n-gram precision, the captions generated by the blind noun translation system are very comparative, even though the system does not do any image processing and merely turns a set of one-hot vectors into the most likely sentence according to the distribution of the training data. Since the commonly used BLEU metric combines the geometric mean of the four precision scores with a brevity penality, this also manifests a shortcoming of the BLEU precision metric. It shows that the evaluation of image captioning systems has to take into account that powerful language models are very good at reproducing the probability distribution of the training data and that the main power of vision is in recognizing basic noun-concepts. The added value of language modeling currently is mostly limited to concept detection as is demonstrated in this otherwise blind experiment. 

\setlength{\tabcolsep}{4pt}
\begin{table}
\begin{center}
\caption{N-gram precision scores for captions generated by a system by Karpathy et al. \cite{karpathy2015deep} compared to a noun translation system, that translates the nouns extracted from the Karpathy et al. image captioning system into captions using a state-of-the-art machine translation system \cite{DBLP:journals/corr/BahdanauCB14}.}
\label{table:headings}
\begin{tabular}{lcccc}
\hline\noalign{\smallskip}
Model & P-1 & P-2 & P-3 & P-4\\
\noalign{\smallskip}
\hline
\noalign{\smallskip}
Image captioning system (Karpathy et al. \cite{karpathy2015deep}) & 63.1 & 35.2 & 17.7 & 9.3\\
Blind noun translation system & 61.9 & 32.0 & 15.7 & 8.2\\
\hline
\end{tabular}
\end{center}
\end{table}
\setlength{\tabcolsep}{1.4pt}

\begin{figure}[t]
\begin{center}
\includegraphics[height=3.5cm]{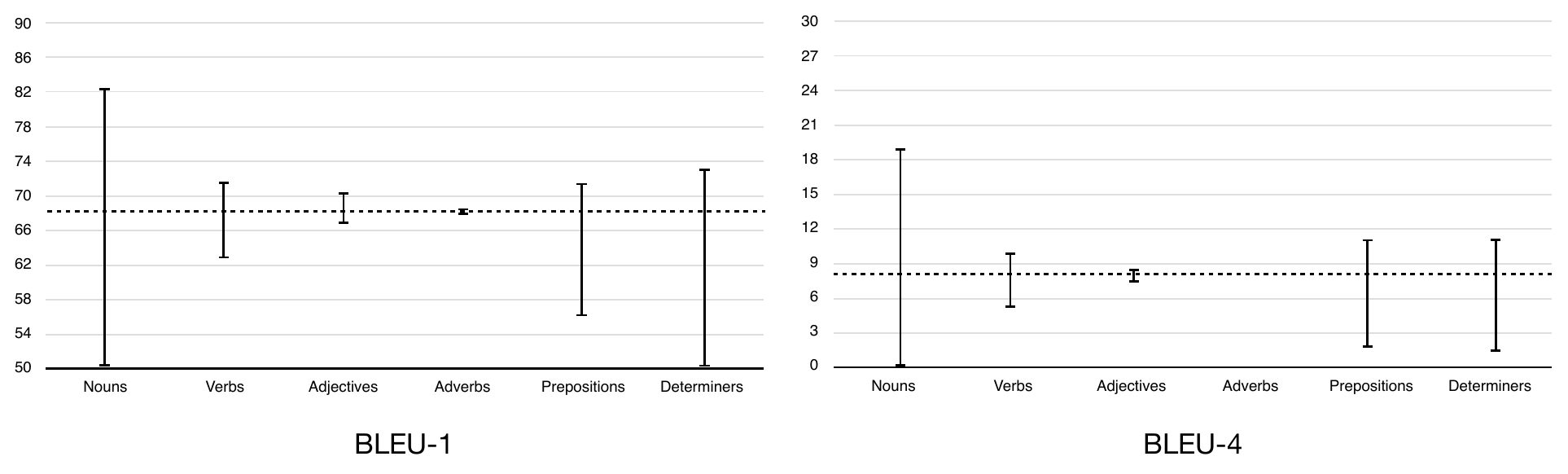}
\end{center}
   \caption{Lower and upper bounds per word category for BLEU-1 (left, ranging from 50 to 90) and 4 (right, ranging from 0 to 30) for the Karpathy et al. \cite{karpathy2015deep} system on the MSCOCO dataset showing possible improvement respectively loss for different word categories.}
\end{figure}

\begin{figure}[t]
\begin{center}
\includegraphics[height=3.5cm]{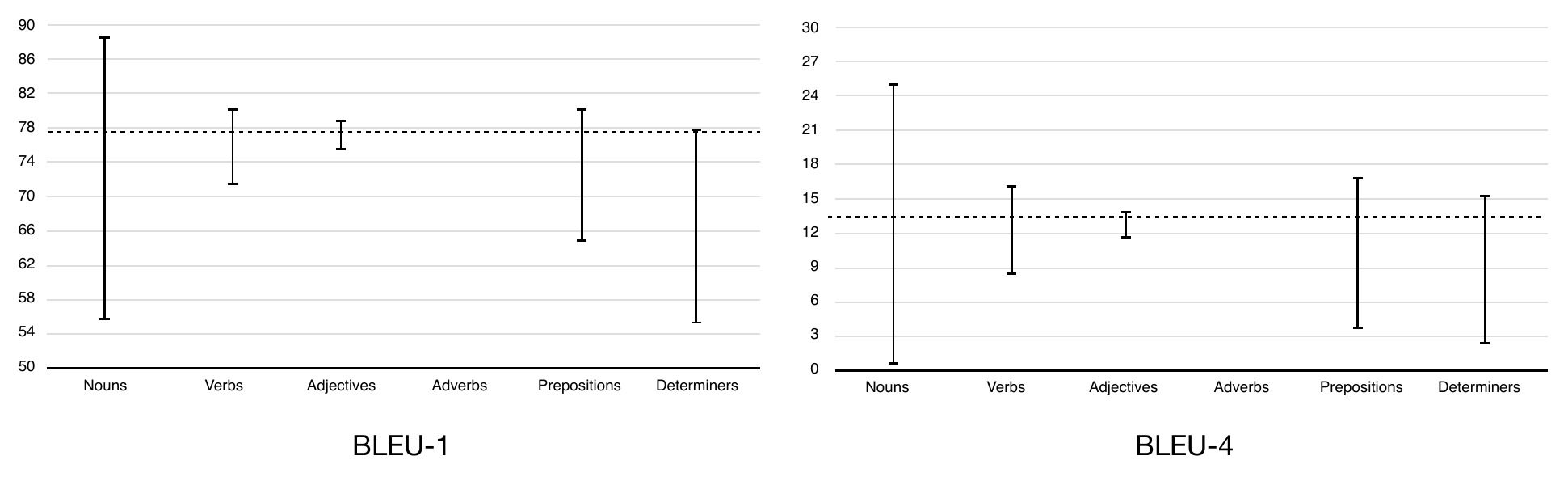}
\end{center}
   \caption{Lower and upper bounds per word category for BLEU-1 (left, ranging from 50 to 90) and 4 (right, ranging from 0 to 30) for the Xu et al. \cite{DBLP:journals/corr/XuBKCCSZB15} system on the MSCOCO dataset showing possible improvement respectively loss for different word categories.}
\end{figure}

The comparison of the contributions of different word categories can be seen in Figures 3 and 4. Each figure visualizes the performance difference regarding (a) unigram precision (BLEU-1) and (b) 4-gram precision (BLEU-4). The dotted line in the middle marks the system performance, i.e., how well the system performs on the MSCOCO Captioning Challenge without any modifications. A line above the system performance highlights the upper bound and indicates how much room for improvement is left, a line below the system performance shows how much worse a system would be performing given that it wouldn't be able to detect any member of the respective category. Figures 3 and 4 show that overall, nouns are currently the most important category where vision makes the most out of the image. Next to nouns, verbs and prepositions have the biggest room for improvement, closely followed by determiners. \\
\textbf{BLEU-1} An analysis of the lower bound shows that the systems already manage to capture a lot of crucial information in the captions. All systems would lose almost 19\% or more in unigram precision (BLEU-1) if they would fail to generate any nouns. Failing to generate determiners would yield a similarly drastic drop in performance. Interestingly, prepositions would also lead to losing 15\% unigram precision performance. The pattern of performance drop is consistent with the three different systems analyzed. \\
At the upper bound, despite the already high performance regarding nouns, there is still a possible improvement of 11,9\% for the Vinyals et al. system (up to BLEU-1 86,8), 11,1\% for the Xu et al. system (up to BLEU-1 88,6), and 14,0\% for the Karpathy et al. system (up to BLEU-1 82,4). Moreover, improvements in capturing and generating verbs, adjectives and prepositions could yield an improvement of up to 3\%  for the BLEU-1 score metric alone. Interestingly, the possible performance gain by improving determiners is low (0.6\%) for the attention-based Xu et al. system and the Vinyals et al. (1.1\%) system and high for the Karpathy system (5\%). Like with the lower bound, the shape of the improvement distribution is consistent between the three systems.  \\
\textbf{BLEU-4} For the lower bound of the 4-gram precision score, failing to generate any nouns eradicates the entire performance of all three systems. A similar pattern can be observed for prepositions and determiners. Adjectives and adverbs are not affected regarding the precision score, which is likely connected to them being largely absent. \\
At the upper bound, an improvement on the generation of nouns could effectively double the performance of the system for the 4-gram precision score. Improving prepositions has considerable potential for all systems, especially for the attention-based system that could improve by 3.3\% through this. Interestingly, even though the number of verbs, and thus 4-gram, per sentence is limited, improving on verbs would yield a similar improvement in the range of 1.7 - 2.6\%.

\section{Discussion}

The results show that a blind noun translation system can generate captions that are comparable to state-of-the-art image captioning systems. This highlights how strong the language modeling capabilities of the LSTM are. It also shows how important it is to critically evaluate the contribution of the LSTM and its intrinsic language modeling capabilities, which should motivate a more rigorous evaluation of image captioning results.  

While we acknowledge the limitations of the BLEU metric for the overall evaluation task, as a precision-based metric, it is still suited to study how much individual word categories contribute to image captioning performance. The results show that it is possible to perform a more qualitative analysis of the contribution of specific linguistic phenomena on the image captioning task performance. The analysis indicates that a considerable improvement in regards to the BLEU precision metric and certain word categories is possible, especially nouns, verbs, and prepositions. 

Image captioning, much like machine translation, needs a reliable automatic metric which can be run quickly at no cost and which correlates with human judgment while taking task-specific challenges into account. To advance from turning the results of an object detection task into a fluent and adequate sentence, a reliable, automatically testable way of evaluating captions is needed. The goal should be to generate meaningful, sharp sentences that convey the semantic content of an image to a variety of stakeholders. 

\bibliographystyle{splncs03}
\bibliography{egbib}
\end{document}